\documentclass[lettersize,journal]{IEEEtran}
\usepackage{amsmath,amsfonts,amssymb}
\usepackage{algorithmic}
\usepackage{algorithm}
\usepackage{array}
\usepackage[caption=false,font=normalsize,labelfont=sf,textfont=sf]{subfig}
\usepackage{textcomp}
\usepackage{stfloats}
\usepackage{url}
\usepackage{verbatim}
\usepackage{graphicx}
\usepackage{cite}
\usepackage{multirow}
\usepackage{booktabs}

\begin{document}

\title{Adaptive Emotional Video Captioning via Affective Heterogeneous Graph Reasoning and Multi-task Joint Learning}

\author{Junbo Wang\IEEEauthorrefmark{1}, Liangyu Fu\IEEEauthorrefmark{1}, Yuke Li\IEEEauthorrefmark{2}, Xuecheng Wu, Zhiyong Wang
\thanks{\IEEEauthorrefmark{2} \textit{Corresponding author: Yuke Li.}}

\thanks{\IEEEauthorrefmark{1} Both authors contributed equally to this work.}

\thanks{Junbo Wang, Liangyu Fu and Yuke Li are with the School of Software, Northwestern Polytechnical University, Xi'an 710129, China (e-mail: jbwang@nwpu.edu.cn; lyfu@mail.nwpu.edu.cn; liyuke@nwpu.edu.cn).}

\thanks{Xuecheng Wu is with the School of Computer Science and Technology, Xi'an Jiaotong University, Xi'an 710049, China (e-mail: wuxc3@stu.xjtu.edu.cn).}

\thanks{Zhiyong Wang is with the School of Computer Science, The University of Sydney, NSW 2006, Australia (e-mail: zhiyong.wang@sydney.edu.au).}

}

\markboth{Journal of \LaTeX\ Class Files,~Vol.~14, No.~8, August~2021}%
{Shell \MakeLowercase{\textit{et al.}}: A Sample Article Using IEEEtran.cls for IEEE Journals}


\maketitle

\begin{abstract}
Emotional video captioning (EVC) aims to describe a video with both factual correctness and affective expressiveness. It requires a model to perceive subtle, ambiguous, and temporally varying emotional cues and translate them into natural language without weakening objective visual content. Existing methods have progressively introduced contextual attention, emotion interpretation, emotion priors, dynamic emotion perception and emotion-cause reasoning. Nevertheless, most of them still depend on either global emotion vectors or rigid hierarchical priors. In recent methods, the tree-structured emotion prior establishes a coarse-to-fine connection between psychological emotion categories and daily emotion words, but its hard subordinate masking may irreversibly suppress correct lexical emotions once the coarse category prediction is inaccurate. It is also limited in representing mixed or overlapping emotions that frequently occur in real videos. To address the issues, we propose SAGML, an adaptive EVC framework via affective heterogeneous graph and multi-task language modeling. Instead of treating the emotion prior as a discrete tree, SAGML constructs a soft affective heterogeneous graph containing catalog-level emotion nodes and lexical-level emotion word nodes. The graph combines category-category affective proximity, category-word corpus co-occurrence, and word-word semantic similarity into a unified topology. Given video features, SAGML first estimates a catalog-level affective distribution and then propagates it over the graph to obtain a topology-driven soft lexical gate. The soft gate is injected into video-to-emotion graph attention as a continuous bias, allowing visually supported lexical emotions to remain recoverable rather than being removed by a hard mask. The resulting affective representation is fed together with visual tokens into a causal language decoder, while dual catalog and lexical heads impose explicit emotion distribution learning on the prompt hidden states. The overall model is trained with a joint objective that combines autoregressive caption generation and emotion distribution supervision. By replacing hard hierarchical pruning with continuous graph reasoning and by coupling generation with multi-level affective prediction, SAGML provides an error-resilient and multi-emotion-aware baseline for EVC.
\end{abstract}


\section{Introduction}
\begin{figure}[t]
    \centering
   \includegraphics[scale=0.68]{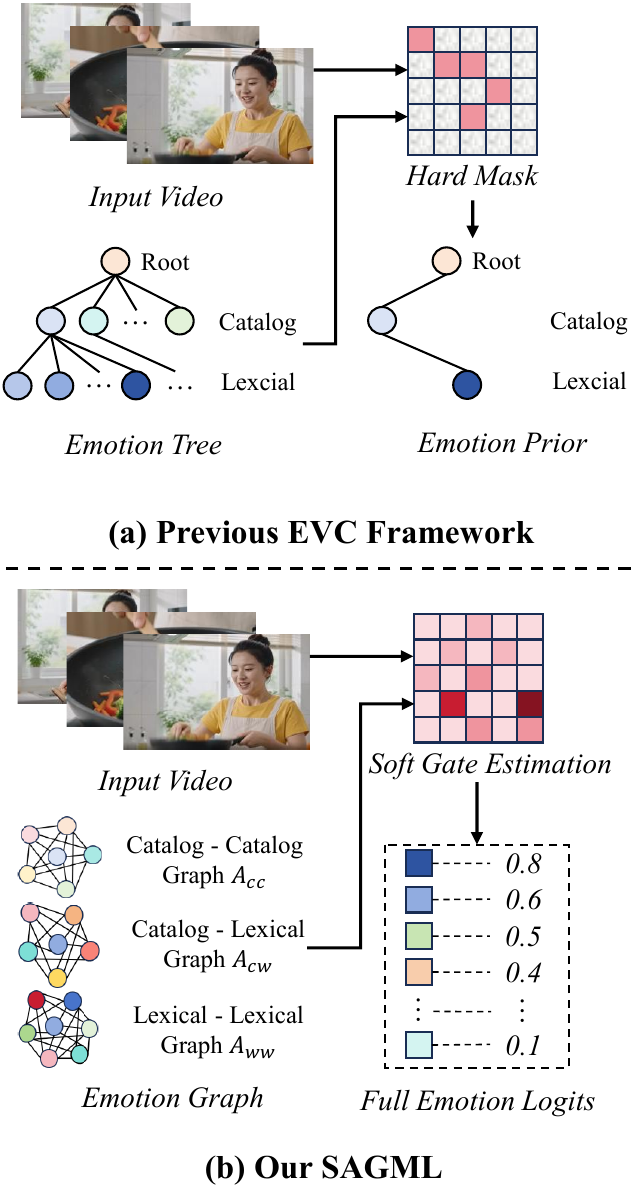}
    \caption{Comparison between the (a) previous EVC framework and (b) our proposed SAGML.}
    \label{fig:intro}
\end{figure}
\IEEEPARstart{V}{ideo} captioning is a fundamental task in multimedia understanding, requiring a model to summarize visual content with fluent natural language. Classical approaches established recurrent convolutional encoders and sequence-to-sequence decoders, and subsequently improved temporal abstraction through hierarchical encoding and frame-level attention~\cite{donahue2015lrcn,venugopalan2015s2vt,pan2016hrne,yao2015describing}. Most classical video captioning systems focus on objective facts, such as objects, actions, scenes, and event transitions. However, user-generated videos shared on social platforms often convey affective intent through facial expressions, body motion, color tone, scene context, and temporal changes. A caption that only reports the factual content may therefore be semantically correct but emotionally incomplete. Emotional video captioning (EVC) addresses this limitation by generating descriptions that jointly preserve factual content and express the intrinsic emotions conveyed by the video.

EVC is challenging for three main reasons. First, emotional cues are usually implicit and ambiguous. The same visual event may evoke different affective interpretations depending on scene context, temporal evolution, and linguistic framing. Second, emotion is not isolated from factual semantics. A caption must describe what happens in the video while selecting emotion words that are grounded in the observed content. Overemphasizing affect can produce emotional but factually drifting captions, whereas insufficient affective modeling leads to bland and neutral descriptions. Third, real videos often contain compound or evolving emotions. A birthday scene may involve joy and surprise, a farewell scene may combine sadness, warmth, and hope. A model that assumes a single dominant emotion may fail to represent such mixed affective states.

Early work on emotion-aware video description introduced fact-emotion dual streams and emotion-oriented datasets, showing that emotional expression is essential for human-like video descriptions~\cite{wang2022emotion}. Contextual attention models further demonstrated that visual and textual contexts should be jointly attended during caption decoding~\cite{song2023contextual}. More recent models explicitly learn video emotion representations. VEIN predicts an emotion distribution over an open psychological vocabulary and coordinates emotion prediction with factual contrastive learning~\cite{song2024vein}. EPAN introduces a perception-priority paradigm in which a model first learns a tree-structured emotion prior and then uses the perceived emotion to guide caption generation~\cite{song2023epan}. This line of work confirms that explicit affective representation is important for EVC.

Despite this progress, rigid hierarchical emotion priors still impose structural constraints that are not fully aligned with the nature of visual emotion. As shown in Fig.~\ref{fig:intro} (a), most work~\cite{han2026heart,ye2025mmecpe,ye2024dcgn,song2023epan} organizes emotions into catalog-level psychological categories and lexical-level daily emotion words, and applies a subordinate masking mechanism to guide coarse-to-fine emotion learning. This design is effective because it narrows the search space for lexical emotion words. However, the binary mask also creates an error cascade: once a coarse category receives a low score, its subordinate lexical words may be blocked even if the visual content supports them. Moreover, a tree encodes mostly one-to-many relations and is less suitable for multi-category words, semantically related emotion words, and co-existing emotions across categories.

Furthermore, another limitation of current EVC models lies in the language decoder. Recurrent or lightweight Transformer decoders can be trained effectively on limited EVC datasets, but they may have insufficient language capacity to express nuanced affective content, especially for low-frequency lexical emotions. Large language models (LLMs) provide strong linguistic priors and compositional generation ability, but directly applying them to EVC can cause a new problem: the model may generate fluent generic sentences while ignoring visual affect. Thus, an EVC decoder should not only receive visual and affective prompts, but also be explicitly regularized to preserve multi-level emotion predictions.

To address these issues, we propose SAGML, a self-adapting emotional video captioning framework based on affective heterogeneous graph reasoning and multi-task language modeling. SAGML replaces the discrete emotion tree with a continuous graph containing two types of nodes: catalog nodes for psychological emotion categories and lexical nodes for daily emotion words. The graph topology integrates three complementary relations: affective proximity between catalog categories, corpus-level conditional association between categories and words, and semantic similarity among lexical emotion words. Based on the catalog distribution inferred from the video, SAGML propagates affective activation over the graph to produce a soft lexical gate. Unlike hard masking, the gate assigns continuous weights and preserves non-zero probability for alternative lexical paths, allowing later visual-emotion attention and language decoding to correct early uncertainty.

SAGML then performs graph-guided affective encoding. Video features query lexical graph nodes, and the soft gate is injected as an attention bias. This makes graph attention both prior-aware and visually adaptive: prior-related lexical nodes are encouraged, while visually matched but initially underweighted nodes can still be selected. The generated affective tokens are concatenated with video tokens and fed into a causal language decoder. To keep the decoder affectively grounded, SAGML attaches dual catalog and lexical classification heads to the prompt hidden states and optimizes them with an emotion distribution learning loss. The final objective jointly trains caption generation and multi-level affective prediction.

The main contributions of this paper are summarized as follows.

\begin{itemize}
    \item We propose SAGML, a self-adapting framework for emotional video captioning that integrates affective graph reasoning with multi-task causal language modeling.
    \item We replace rigid tree-structured emotion masking with a unified continuous affective heterogeneous graph and topology-driven soft gating with graph-biased video-to-emotion attention, jointly modeling category–category, category–word, and word–word relations to enable error-resilient lexical emotion selection and better represent mixed emotional states.
    \item We introduce a prompt-level multi-task learning strategy for EVC, where catalog and lexical emotion distribution heads regularize the language decoder jointly with autoregressive caption generation.
    \item We fine-tuned a large language model on the EVC task for the first time, equipping it with more precise emotion understanding capabilities.
    \item Experiments on EmVidCap-S, EmVidCap-L, and EmVidCap show consistent improvements over existing methods. On the full EmVidCap dataset, SAGML surpasses the strongest comparison method by 6.7 points in Acc$_{sw}$, 7.7 points in Acc$_c$, 9.0 points in CIDEr, and 8.6 points in CFS.
\end{itemize}

\section{Related Work}

\subsection{Factual and Stylized Visual Captioning}
Video captioning translates a temporally ordered visual signal into a natural-language description. Early neural systems established the encoder--decoder formulation by coupling convolutional video representations with recurrent language models. LRCN connected convolutional visual features to recurrent sequence models for end-to-end visual recognition and description~\cite{donahue2015lrcn}. S2VT then used a sequence-to-sequence architecture to encode video frames and decode words~\cite{venugopalan2015s2vt}, while HRNE hierarchically summarized short frame subsequences to model longer temporal transitions~\cite{pan2016hrne}. Temporal attention further allowed the decoder to select different video segments for different output words~\cite{yao2015describing}. These studies established temporal representation learning as a central component of factual video description.

Later work improved factual grounding and long-range correspondence in several ways. M3 introduced a shared visual--textual memory to model long-term multimodal dependencies and guide attention~\cite{wang2018m3}. RecNet added a backward objective that reconstructs video features from decoder states~\cite{wang2018recnet}. SGN groups frames according to discriminative phrases in the partially generated caption~\cite{ryu2021sgn}, and SwinBERT performs end-to-end spatio-temporal encoding and caption generation with a learned sparse attention mask~\cite{lin2022swinbert}. Beyond single-sentence description, dense event captioning jointly localizes and describes multiple temporally situated events in a video~\cite{krishna2017dense}. This line of work improves the representation of objects, actions, event boundaries, and temporal dependencies, but its learning objectives do not explicitly require the generated language to convey the affect expressed by a video.

Research on visual sentiment and stylized captioning provides an earlier connection between visual grounding and non-factual language. SentiBank constructed a large-scale visual sentiment ontology from adjective--noun pairs and trained detectors for the resulting affective concepts~\cite{borth2013sentibank}. Building on this form of visual sentiment representation, SentiCap combines factual and sentiment-specific recurrent streams to generate positive or negative image descriptions~\cite{mathews2016senticap}. StyleNet factors recurrent parameters to learn controllable styles from factual image--caption pairs and unpaired stylized text~\cite{gan2017stylenet}. Chen \emph{et al.} further introduced a style-factual LSTM and adaptive learning to balance visual fidelity with a requested linguistic style~\cite{chen2018factual}. These studies show that affective wording should be introduced without discarding visual content. However, a prescribed style or polarity is different from EVC: in EVC, the emotion must first be inferred from the video and then expressed with visually appropriate words.

\subsection{Emotional Video Captioning}
Wang et al. established the main EVC benchmark by constructing EmVidCap-S from rewritten MSVD captions, annotating the longer EmVidCap-L subset from VideoEmotion-8, and combining them as EmVidCap~\cite{wang2022emotion}. Their Fact Transfer model uses separate factual and emotional streams and fuses the two word distributions during decoding. CANet replaces this fixed two-stream fusion with a unified model that attends to both video features and previously generated words, and introduces video--caption contrastive learning to improve contextual representations~\cite{song2023contextual}. These methods demonstrate the benefit of affective supervision, although emotion is still learned mainly through caption-level mapping rather than an explicit structured representation.

Subsequent work makes emotion perception more explicit. VEIN predicts a distribution over a 179-word emotion vocabulary, aggregates the most responsive words into an emotion vector, and coordinates emotional indication with factual contrastive learning~\cite{song2024vein}. EPAN organizes 34 psychological categories and 179 lexical emotion words into a two-level tree. It predicts catalog emotions first and applies a subordinate mask before lexical emotion encoding, thereby implementing coarse-to-fine emotion perception~\cite{song2023epan}. This hierarchy improves interpretability and emotion accuracy, but its binary top-$K$ selection can suppress a useful lexical word when the preceding catalog prediction is incorrect.

More recent EVC models focus on temporal variation, cross-modal grounding, and fine-grained causes. DCGN updates emotion representations at each decoding step through element- and subspace-level evolution, and estimates emotion intensity before injecting affective features into the decoder~\cite{ye2024dcgn}. ECPA targets human-centric captions with emotion-recognition and facial-action-unit visual prompts, sentence- and word-level textual prompts, and corresponding cross-modal alignment objectives~\cite{wang2025ecpa}. MM-ECPE treats emotional words and their visual causes as paired evidence: it refines visual and lexical features in two rounds, aligns the final emotion--cause representations contrastively, and conditions a pretrained decoder on the resulting pairs~\cite{ye2025mmecpe}. HEART complements category-level emotion learning with entity-, action-, and event-level semantic extraction and a temporal pyramid, and introduces EmoStruct with subject- and predicate-oriented emotion annotations~\cite{han2026heart}. Collectively, these methods reduce different sources of EVC error, including static emotion modeling, insufficient facial detail, missing causes, and incomplete temporal semantics. SAGML addresses a separate but related issue: it changes the emotion prior itself from a hard hierarchy into a weighted heterogeneous graph, so uncertainty can be propagated rather than irreversibly pruned.

\subsection{Visual Emotion Understanding and Structured Priors}
Visual emotion analysis studies affective states conveyed or elicited by visual content. For person-centric images, EMOTIC demonstrated that body appearance and surrounding scene context provide complementary evidence and represented emotion with both discrete categories and continuous valence--arousal--dominance dimensions~\cite{kosti2020emotic}. VideoEmotion-8 showed that appearance, audio, and semantic attributes likewise provide complementary evidence for emotion recognition in user-generated videos~\cite{jiang2014predicting}. VAANet later integrated spatial, channel, and temporal attention in an end-to-end audio-visual model~\cite{zhao2020vaanet}. Affect2MM modeled time-varying emotion with facial, scene, aesthetic, action, and script cues, together with temporal causality~\cite{mittal2021affect2mm}. Large-scale web supervision has also been used to learn transferable visual emotion representations over a substantially richer emotion vocabulary~\cite{wei2020emotion}. These recognition models provide useful affective features, but EVC additionally requires the selected emotion to be grounded in a factual sentence.

Emotion is ambiguous and often multi-label, which motivates distributional and structured prediction. The Plutchik model organizes primary emotions, intensity variants, and compound relations~\cite{plutchik1980emotion}, and it also underlies the 34-category lexicon used by EmVidCap and later EVC models. Image emotion distribution learning has used graph convolution to model correlations among emotion labels instead of predicting each label independently~\cite{he2019imageedl}. Circular-structured emotion distribution learning further encodes polarity, type, and intensity~\cite{yang2021circular}. EPAN similarly shows that catalog--lexical structure can regularize EVC~\cite{song2023epan}, yet a tree represents only fixed parent-child edges.

Graph neural networks provide a more flexible basis for structured affective reasoning. Graph convolutional networks propagate information through local neighborhoods to jointly encode node attributes and topology~\cite{kipf2017gcn}, while relational graph convolution assigns relation-specific transformations to heterogeneous edges~\cite{schlichtkrull2018rgcn}. Graph attention additionally learns different relevance weights for connected nodes~\cite{velickovic2018gat}. These properties are useful for emotion vocabularies because category proximity, category--word association, and word--word similarity are heterogeneous and many-to-many. SAGML combines these relations in one topology and converts the predicted catalog distribution into a continuous lexical gate. The design retains a psychological prior while allowing visually supported alternatives and cross-category lexical relations to remain available.

\subsection{Foundation Models and Parameter-Efficient Generation}
Vision-language pretraining supplies representations that are better aligned with text than conventional classification features. CLIP learns transferable image representations by contrastively matching images and natural-language descriptions~\cite{radford2021clip}. Frozen showed that a visual encoder can map images into continuous prefix embeddings understood by a pretrained frozen language model~\cite{tsimpoukelli2021frozen}. Flamingo subsequently connected pretrained vision and language components with gated cross-attention and supported interleaved image, video, and text inputs~\cite{alayrac2022flamingo}. BLIP-2 uses a lightweight querying transformer to bridge frozen visual encoders and frozen language models~\cite{li2023blip2}. Together, these models establish continuous visual prompting as an efficient alternative to training a multimodal generator from scratch. On the language side, Qwen2.5 provides a strong autoregressive prior for compositional generation~\cite{qwen2025qwen25technicalreport}, while LoRA adapts large language models through low-rank updates instead of full parameter tuning~\cite{hu2022lora}.

These advances are relevant to EVC but do not by themselves ensure emotional grounding. A pretrained decoder may produce fluent captions while underusing subtle video affect, especially when emotional words are sparse in the task-specific corpus. SAGML therefore uses the language model in three coupled roles. First, its embedding space initializes typed catalog and lexical nodes. Second, projected visual and graph-derived affective tokens form the continuous prefix for caption generation. Third, the pooled prefix hidden state is supervised by catalog- and lexical-level distribution objectives. The decoder is thus adapted not only to predict the next token, but also to preserve the multi-level affective evidence supplied by the graph.

\section{Proposed Method}

\begin{figure*}[t]
    \centering
    \begin{center}
   \includegraphics[width=\linewidth]{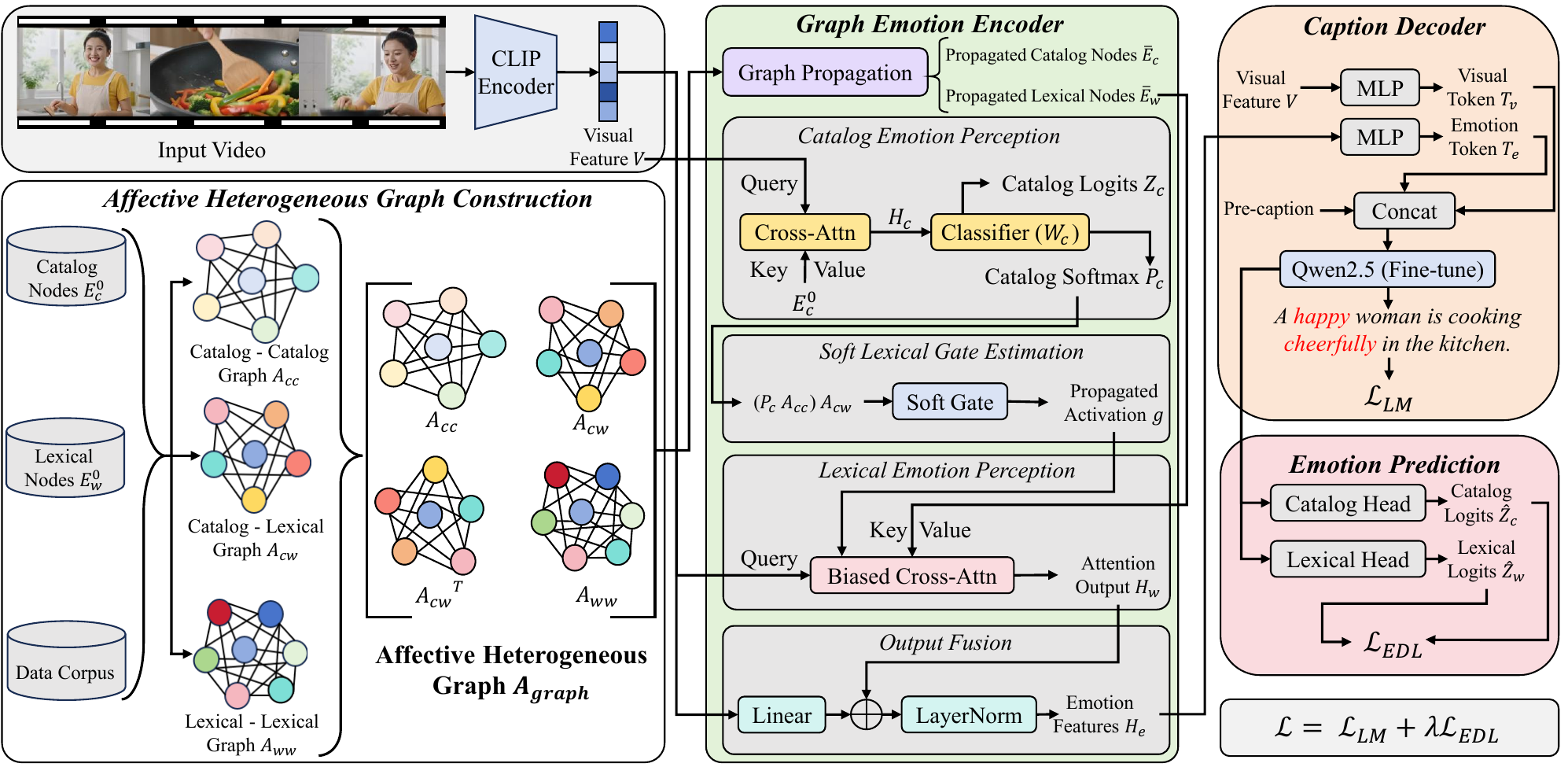}
    \end{center}
    \caption{Overview of SAGML. Pre-extracted CLIP features are contextualized as visual tokens $V$. Typed catalog and lexical nodes form an affective heterogeneous graph with catalog--catalog, catalog--lexical, and lexical--lexical relations. Catalog perception produces $P_c$, which is propagated through $A_{cc}$ and $A_{cw}$ to obtain the soft lexical gate $g$. The gate biases visual-to-lexical attention, and the resulting $H_w$ is fused with a visual residual to produce $H_e$. Projected visual and emotion tokens condition a LoRA-adapted Qwen decoder. Catalog and lexical heads on the pooled prompt states are trained jointly with caption generation.}
    \label{fig:overall}
\end{figure*}

\subsection{Task Definition and Overview}
Given a video $X$, emotional video captioning generates a sentence
$Y=(y_1,\ldots,y_T)$ that describes the observed content and expresses the
affect conveyed by the video. We use a two-level emotion vocabulary:
$\mathcal{C}=\{c_i\}_{i=1}^{M_c}$ contains catalog emotions and
$\mathcal{W}=\{w_j\}_{j=1}^{M_w}$ contains lexical emotion words. Following
EmVidCap, $M_c=34$ and $M_w=179$.

As illustrated in Fig.~\ref{fig:overall}, SAGML contains three trainable
stages. First, a lightweight visual encoder contextualizes frame-level CLIP
features. Second, a graph emotion encoder relates catalog and lexical emotions,
infers a catalog distribution from the video, and converts it into a continuous
lexical prior for visual-to-emotion attention. Third, projected visual and
emotion sequences form a continuous prefix for a Qwen-based decoder. Two
emotion heads attached to the Qwen prompt states provide catalog- and
lexical-level distribution supervision. All trainable components are optimized
end to end with language-modeling and emotion-distribution losses; the CLIP
features themselves are extracted offline.

\subsection{Visual Encoding}
We uniformly sample $N$ frames and use a frozen CLIP ViT-B/32 encoder
to extract appearance features~\cite{radford2021clip}:
\begin{equation}
F=[f_1;\ldots;f_N]\in\mathbb{R}^{N\times d_f}.
\end{equation}
The implementation uses $N=30$ and $d_f=512$. A learned affine projection
maps the features to dimension $D$, after which a one-layer, one-head
self-attention encoder contextualizes the frame sequence:
\begin{equation}
V=\operatorname{Enc}_{vis}(FW_v+b_v)
  \in\mathbb{R}^{N\times D}.
\end{equation}
Here $D=300$; the self-attention key and value dimensions are both 32,
and the position-wise feed-forward dimension is 512. No motion stream is
consumed by the current visual encoder. The resulting $V$ is used as the query
sequence in both levels of emotion perception and is also passed directly to
the caption decoder.

\subsection{Qwen-Aligned Affective Node Initialization}
The graph operates in the hidden space of the selected Qwen model. Let
$H$ denote its hidden dimension, which is 3584 for the default Qwen2.5-Instruct
backbone~\cite{qwen2025qwen25technicalreport}. To distinguish the two node types, each
emotion string $s$ is converted into a typed phrase:
\begin{equation}
\rho_c(s)=\text{``emotion category: ''}\Vert s,
\end{equation}

\begin{equation}
    \rho_w(s)=\text{``emotion word: ''}\Vert s,
\end{equation}

where $\Vert$ denotes string concatenation. If
$\operatorname{Tok}(\rho(s))=(u_1,\ldots,u_L)$ and
$E_{\mathrm{LLM}}$ is the Qwen input embedding table, the node feature is
\begin{equation}
e(s)=\frac{1}{L}\sum_{\ell=1}^{L}E_{\mathrm{LLM}}(u_\ell)
     \in\mathbb{R}^{H}.
\end{equation}
Applying this operation to all catalog and lexical strings yields
$E_c^0\in\mathbb{R}^{M_c\times H}$ and
$E_w^0\in\mathbb{R}^{M_w\times H}$, respectively. These features are
computed from the same embedding table used by the caption decoder. GloVe
embeddings~\cite{pennington2014glove} are retained only for constructing one
component of the lexical adjacency; they are not used as Qwen token
embeddings.

\subsection{Affective Heterogeneous Graph Construction}
SAGML represents the two emotion levels as an affective heterogeneous graph
\begin{equation}
\mathcal{G}=(\mathcal{V},\mathcal{E}),\qquad
\mathcal{V}=\mathcal{V}_c\cup\mathcal{V}_w.
\end{equation}
For a nonnegative matrix $B$, define row normalization as
\begin{equation}
\operatorname{RN}(B)_{ij}=
\frac{B_{ij}}{\max(\sum_k B_{ik},\epsilon)}.
\end{equation}

\textit{Catalog--catalog relation.}
The implementation imposes a smooth cyclic neighborhood over the fixed catalog
index order. For indices $i$ and $j$,
\begin{equation}
d_{\mathrm{cyc}}(i,j)=
\min\bigl(|i-j|,M_c-|i-j|\bigr),
\end{equation}
\begin{equation}
A_{cc}=\operatorname{RN}\left(
\left[\exp\left(-\frac{d_{\mathrm{cyc}}(i,j)^2}{2\sigma^2}\right)
\right]_{ij}\right),
\end{equation}
with $\sigma=1.5$. Thus catalog activation can be smoothed to neighboring
indices rather than remaining one-hot.

\textit{Catalog--lexical relation.}
For each training caption, the loader extracts the unique words that occur in
$\mathcal{W}$ and maps each word to one or more catalog categories using the
provided emotion lexicon. Let $n_{ij}^{cw}$ count the resulting association
between $c_i$ and $w_j$, and let $n_i^c=\sum_j n_{ij}^{cw}$. Additive
smoothing gives
\begin{equation}
\widetilde A_{cw}[i,j]=
\frac{n_{ij}^{cw}+\alpha}{n_i^c+\alpha M_w},\qquad
A_{cw}=\operatorname{RN}(\widetilde A_{cw}),
\end{equation}
where $\alpha=0.1$. This construction retains multi-parent words and assigns a
nonzero smoothed association to every catalog--lexical pair.

\textit{Lexical--lexical relation.}
The lexical adjacency combines Qwen similarity, GloVe similarity, and
caption-level co-occurrence. For a feature matrix $E$, the thresholded
nonnegative cosine matrix is
\begin{equation}
S_{\tau}(E)[i,j]=
\begin{cases}
\max(\operatorname{cos}(E_i,E_j),0),
& \operatorname{cos}(E_i,E_j)\geq\tau,\\
0,&\text{otherwise},
\end{cases}
\end{equation}
with unit diagonal. Let $G_w$ be the 300-dimensional GloVe lexical features.
For the corpus term, $n_{ij}^{ww}$ counts ordered pairs of unique emotion
words co-occurring in a caption and $n_i^w$ counts captions containing $w_i$:
\begin{equation}
C_{ww}[i,j]=
\frac{n_{ij}^{ww}+\alpha}{n_i^w+\alpha M_w}.
\end{equation}
After setting the diagonal of $C_{ww}$ to one and row-normalizing it, the
three sources are fused as
\begin{equation}
A_{ww}=\operatorname{RN}\left(
\beta_L S_\tau(E_w^0)+
\beta_G S_\tau(G_w)+
\beta_R\operatorname{RN}(C_{ww})\right).
\end{equation}
The implementation uses $\tau=0.5$ and
$(\beta_L,\beta_G,\beta_R)=(0.5,0.3,0.2)$. Before the final row
normalization, the diagonal is reset to one.

The complete adjacency is
\begin{equation}
A_{\mathrm{graph}}=\operatorname{RN}\left(
\begin{bmatrix}
A_{cc}&A_{cw}\\
A_{cw}^{\top}&A_{ww}
\end{bmatrix}\right).
\end{equation}
One propagation step followed by a two-layer projection produces
\begin{equation}
\begin{bmatrix}\bar E_c\\\bar E_w\end{bmatrix}
=\Phi\left(
A_{\mathrm{graph}}
\begin{bmatrix}E_c^0\\E_w^0\end{bmatrix}\right),
\end{equation}

\begin{equation}
    \Phi(x)=\operatorname{LN}\bigl(
W_2\operatorname{GELU}(W_1x+b_1)+b_2\bigr).
\end{equation}

The current implementation uses $\bar E_w$ in lexical perception. Although
$\bar E_c$ is computed by the propagation routine, catalog perception uses
the initial catalog nodes $E_c^0$.

\subsection{Catalog Emotion Perception}
The catalog branch uses each visual token as a query and the initial catalog
nodes as keys and values:
\begin{equation}
Q_c=VW_Q^c,\qquad
K_c=E_c^0W_K^c,\qquad
U_c=E_c^0W_V^c.
\end{equation}
The attention matrix and catalog-aware visual sequence are
\begin{equation}
A_c=\operatorname{softmax}\left(
\frac{Q_cK_c^\top}{\sqrt H}\right),
\end{equation}

\begin{equation}
    H_c=A_cU_c.
\end{equation}

Mean pooling over the $N$ visual positions gives the encoder-side catalog
logits and probability distribution shown in Fig.~\ref{fig:overall}:
\begin{equation}
Z_c=\operatorname{Linear}_c\left(
\frac{1}{N}\sum_{n=1}^{N}H_c[n]\right),
\end{equation}

\begin{equation}
    P_c=\operatorname{softmax}(Z_c).
\end{equation}

No top-$K$ decision is applied. Consequently, gradients from the downstream
soft gate can reach the catalog branch even though $Z_c$ is not assigned a
separate encoder-side classification loss.

\subsection{Topology-Driven Soft Lexical Gate}
The catalog distribution is first smoothed over $A_{cc}$ and then transferred
to lexical nodes through $A_{cw}$:
\begin{equation}
r_w=(P_cA_{cc})A_{cw}\in\mathbb{R}^{M_w}.
\end{equation}
The propagated activation is transformed by an $M_w$-to-$M_w$ linear layer:
\begin{equation}
g=\max\left(
\sigma\bigl(\operatorname{Linear}_g(r_w)\bigr),\epsilon\right),
\end{equation}
where the maximum is element-wise and $\epsilon=10^{-6}$. The gate layer is
initialized as the identity with zero bias. Unlike a binary subordinate mask,
$g_j>0$ for every lexical node, so the catalog prior can attenuate a word
without making it unreachable.

\subsection{Graph-Biased Lexical Affective Attention}
The lexical branch attends from visual tokens to the propagated lexical nodes:
\begin{equation}
Q_w=VW_Q^w,\qquad
K_w=\bar E_wW_K^w,\qquad
U_w=\bar E_wW_V^w.
\end{equation}

The logarithm of the gate is broadcast to all visual positions and added to
the scaled dot-product scores:
\begin{equation}
A_w=\operatorname{softmax}\left(
\frac{Q_wK_w^\top}{\sqrt H}
+\mathbf{1}_N\log(g)^\top\right),
\end{equation}

\begin{equation}
    H_w=A_wU_w.
\end{equation}

Thus $g$ behaves as a multiplicative prior after the softmax, while the
visual--lexical compatibility term remains able to compensate for a weak
prior. The attention output is fused with a learned visual residual:
\begin{equation}
H_e=\operatorname{LayerNorm}\left(
H_w+\operatorname{Linear}_r(V)\right).
\end{equation}
The implementation additionally computes lexical diagnostic logits by
classifying the temporal mean of $H_e$. These encoder-side lexical logits are
returned for analysis but are not included in the current loss. The supervised
catalog and lexical heads described below operate on Qwen prompt states.

\subsection{Qwen-Based Prompt Caption Decoder}
Two independent MLPs map the visual and affective sequences to the Qwen
hidden dimension:
\begin{equation}
T_v=\operatorname{MLP}_v(V),\qquad
T_e=\operatorname{MLP}_e(H_e),
\end{equation}
where each projector is Linear--GELU--Linear--LayerNorm and
$T_v,T_e\in\mathbb{R}^{N\times H}$. The caption tensor produced by the data
loader is converted back to text and re-tokenized with the Qwen tokenizer.
An EOS token is appended, and its input embeddings are denoted by $T_y$.
During teacher-forced training, the decoder input is
\begin{equation}
\mathcal{T}=[T_v;T_e;T_y].
\end{equation}
The first $2N$ positions form the continuous multimodal prompt. Their labels
are set to $-100$, so the native causal-language-model loss is evaluated only
on caption tokens. Accordingly,
\begin{equation}
p(y_t\mid y_{<t},X)=
\operatorname{Qwen}\bigl(y_t\mid[T_v;T_e],y_{<t}\bigr).
\end{equation}
The ``Pre-caption'' path in Fig.~\ref{fig:overall} denotes $y_{<t}$: ground
truth caption prefixes are used during training, whereas previously generated
tokens are used at inference. Generation starts from the Qwen BOS token and
uses greedy next-token selection until EOS or the maximum length is reached.

The Qwen2.5-Instruct backbone is not fully fine-tuned. LoRA~\cite{hu2022lora} is
applied only to the query and value projections of its attention layers, with
rank 8, scaling factor 16, and dropout 0.05. The remaining base-model
parameters stay frozen, while the two prompt projectors, graph encoder, visual
encoder, and prediction heads are trainable. Gradient checkpointing is enabled
for the Qwen backbone.

\subsection{Prompt-Level Multi-Task Emotion Prediction}
Let $H^{p}\in\mathbb{R}^{2N\times H}$ denote the final-layer Qwen hidden
states at the visual and emotion prompt positions. Their mean is
\begin{equation}
h_{\mathrm{vibe}}=\frac{1}{2N}
\sum_{i=1}^{2N}H^{p}[i].
\end{equation}
The catalog and lexical heads in Fig.~\ref{fig:overall} produce
\begin{equation}
\hat z_c=\operatorname{Linear}_{c}^{dec}(h_{\mathrm{vibe}}),
\qquad
\hat z_w=\operatorname{Linear}_{w}^{dec}(h_{\mathrm{vibe}}).
\end{equation}
Because the heads read the contextualized prompt rather than generated words,
their supervision directly constrains how visual and graph-derived emotion
evidence is represented inside the decoder.

\subsection{Training Objective}
The captioning term is the Qwen autoregressive loss:
\begin{equation}
\mathcal{L}_{LM}=-\sum_{t=1}^{T}\log p(y_t\mid y_{<t},X).
\end{equation}
Catalog and lexical labels are obtained from emotion words occurring in each
training caption and their lexicon-defined catalog mappings. The loader retains
up to three labels at each level. The target constructor counts valid indices
and normalizes their counts to form distributions
$q_c\in\mathbb{R}^{M_c}$ and $q_w\in\mathbb{R}^{M_w}$; samples without a
valid target at a level are excluded from that level's loss. The
emotion-distribution loss is
\begin{equation}
\mathcal{L}_{EDL}=
D_{\mathrm{KL}}\left(q_c\parallel
\operatorname{softmax}(\hat z_c)\right)
+D_{\mathrm{KL}}\left(q_w\parallel
\operatorname{softmax}(\hat z_w)\right).
\end{equation}
The implemented total loss is
\begin{equation}
\mathcal{L}=\mathcal{L}_{LM}+\lambda\mathcal{L}_{EDL}.
\end{equation}

\subsection{Discussion}
The main distinction between SAGML and a hard catalog-to-word tree lies in
how the prior enters lexical selection. A hard mask makes the lexical candidate
set a discrete consequence of the catalog decision. SAGML instead maps $P_c$
to the strictly positive gate $g$ and adds $\log g$ to visual--lexical
attention. The graph therefore changes the relative preference among emotion
words without deleting alternatives. The prompt-level objectives then require
the LoRA-adapted decoder to retain this affective evidence while learning the
caption likelihood.

The current Qwen-aligned implementation also reduces the representation mismatch between affective reasoning and language generation. Catalog and lexical nodes are initialized with Qwen input embeddings, visual and affective tokens are projected into the same hidden dimension, and prompt-level heads supervise the affective content encoded in the Qwen prefix. GloVe remains useful, but only as an auxiliary word-word relation prior inside the graph topology. As a result, the model keeps the interpretability of structured emotion priors while using the linguistic capacity of a large causal language model for final caption generation.

\section{Experiments}

\subsection{Datasets}
We evaluate SAGML on the three standard splits introduced with EmVidCap~\cite{wang2022emotion}. They differ substantially in video source, caption length, and annotation protocol, and therefore test complementary aspects of emotional captioning.

\textbf{EmVidCap-S} is derived from MSVD~\cite{chen2011msvd}. Annotators rewrote factual MSVD descriptions by inserting visually compatible emotion expressions. It contains 374 videos: 240 videos with 8,169 emotional captions for training and 134 videos with 4,611 captions for testing. Its clips and sentences are relatively short, so the split mainly evaluates fine-grained emotion-word selection under otherwise familiar factual content.

\textbf{EmVidCap-L} is constructed from the VideoEmotion-8 benchmark~\cite{jiang2014predicting}. Its captions are written from scratch rather than obtained by modifying factual references. The split contains 1,141 training videos with 19,398 captions and 382 test videos with 6,527 captions. Compared with EmVidCap-S, it has longer videos, longer sentences, and more varied emotional phrasing, making temporal content selection and factual--affective coordination more difficult.

\textbf{EmVidCap} combines the two subsets while retaining their official splits. It contains 1,381 training videos with 27,567 captions and 516 test videos with 11,138 captions. We use this combined benchmark for all ablation and parameter studies because it exposes the model to both rewritten short captions and freely annotated long descriptions.

\subsection{Evaluation Metrics}
We report semantic, emotional, and hybrid measures. Semantic quality is evaluated with BLEU-1 to BLEU-4~\cite{papineni2002bleu}, METEOR~\cite{banerjee2005meteor}, ROUGE-L~\cite{lin2004rouge}, and CIDEr~\cite{vedantam2015cider}. BLEU measures modified $n$-gram precision, METEOR additionally considers flexible word matching, ROUGE-L is based on the longest common subsequence, and CIDEr weights consensus with multiple human references.

Following the EmVidCap protocol~\cite{wang2022emotion}, emotional correctness is measured by emotion-word accuracy Acc$_{sw}$ and emotion-sentence accuracy Acc$_c$. Let $N_r$ and $N_w$ be the numbers of correct and incorrect generated emotion words, $\Phi$ the set of generated captions without an emotion word, $N_r^*$ the number of captions containing at least one correct emotion word, $N_h$ the number containing both correct and incorrect emotion words, and $\mathcal{S}$ the complete generated-caption set. The two scores are
\begin{equation}
\mathrm{Acc}_{sw}=\frac{N_r}{N_r+N_w+|\Phi|},
\end{equation}

\begin{equation}
\mathrm{Acc}_c=\exp\left(1-\frac{N_r^*}{N_r^*-N_h}\right)\frac{N_r^*}{|\mathcal{S}|}.
\end{equation}

We further report BFS and CFS to avoid evaluating factual and emotional quality in isolation:

\begin{equation}
\mathrm{BFS} = \beta\sum_{n=1}^{4}\omega_n\mathrm{BLEU}_n+ (1-\beta)\overline{\mathrm{Acc}},
\end{equation}

\begin{equation}
    \mathrm{CFS} = \beta\,\mathrm{CIDEr} + (1-\beta)\overline{\mathrm{Acc}},
\end{equation}

where $\overline{\mathrm{Acc}}=(\mathrm{Acc}_{sw}+\mathrm{Acc}_c)/2$, $\beta=0.8$, and $(\omega_1,\omega_2,\omega_3,\omega_4)=(0.1,0.2,0.3,0.4)$. All methods use the same 34-category/179-word lexicon and evaluation implementation.

\begin{table*}[t]
\centering
\caption{Performance comparison on EmVidCap-S, EmVidCap-L, and EmVidCap.}
\label{tab:main_evc}
\begin{tabular}{llcc|ccccccc|cc}
\toprule
Dataset & Method & Acc$_{sw}$ & Acc$_c$ & B-1 & B-2 & B-3 & B-4 & METEOR & ROUGE-L & CIDEr & BFS & CFS \\
\midrule
\multirow{9}{*}{EmVidCap-S}
& SA-LSTM~\cite{wang2018recnet} & 68.8 & 67.2 & 80.7 & 67.9 & 56.3 & 45.5 & 33.0 & 68.2 & 72.1 & 59.0 & 71.3 \\
& SGN~\cite{ryu2021sgn} & 73.9 & 73.1 & 77.5 & 62.7 & 51.3 & 41.1 & 30.6 & 63.6 & 71.0 & 56.4 & 71.5 \\
& FT~\cite{wang2022emotion} & 69.4 & 67.1 & 77.2 & 60.3 & 47.4 & 36.3 & 29.0 & 63.4 & 62.5 & 52.5 & 63.7 \\
& CANet~\cite{song2023contextual} & 78.7 & 76.8 & 78.5 & 64.0 & 52.1 & 41.8 & 30.8 & 65.7 & 74.4 & 57.9 & 75.1 \\
& VEIN~\cite{song2024vein} & 82.7 & 82.1 & 82.0 & 68.4 & 57.1 & 45.9 & 33.0 & 69.0 & 79.6 & 62.4 & 80.2 \\
& EPAN~\cite{song2023epan} & 84.1 & 82.8 & 82.5 & 69.6 & 57.8 & 46.2 & 34.4 & 69.8 & 80.6 & 63.1 & 81.1 \\
& DCGN~\cite{ye2024dcgn} & 86.5 & 85.7 & 84.5 & 70.9 & 59.2 & 48.7 & 35.7 & 71.0 & 85.2 & 65.7 & 86.6 \\
& MM-ECPE~\cite{ye2025mmecpe} & 88.7 & 87.6 & 86.9 & 72.7 & 60.5 & 49.6 & 38.1 & 74.6 & 88.3 & 66.6 & 88.3 \\
& \textbf{SAGML} & \textbf{93.5} & \textbf{92.3} & \textbf{93.6} & \textbf{78.5} & \textbf{65.4} & \textbf{53.8} & \textbf{40.8} & \textbf{77.2} & \textbf{96.1} & \textbf{71.5} & \textbf{95.5} \\
\midrule
\midrule
\multirow{8}{*}{EmVidCap-L}
& SA-LSTM~\cite{wang2018recnet}  & 48.6 & 47.1 & 71.0 & 51.1 & 34.5 & 22.5 & 19.6 & 40.7 & 30.2 & 38.9 & 33.7 \\
& CANet~\cite{song2023contextual} & 41.9 & 39.7 & 66.9 & 44.8 & 29.3 & 19.3 & 18.2 & 37.9 & 23.3 & 33.9 & 26.8 \\
& VEIN~\cite{song2024vein} & 57.4 & 56.8 & 71.6 & 52.1 & 37.4 & 26.3 & 20.9 & 41.7 & 33.4 & 43.0 & 39.2 \\
& EPAN~\cite{song2023epan} & 63.8 & 62.3 & 73.6 & 54.0 & 38.3 & 27.0 & 21.2 & 42.3 & 34.7 & 45.0 & 40.4 \\
& DCGN~\cite{ye2024dcgn} & 71.0 & 69.4 & 74.5 & 55.3 & 40.0 & 28.1 & 23.4 & 47.7 & 41.5 & 47.3 & 46.9 \\
& HEART~\cite{han2026heart} & 55.8 & 54.2 & 71.5 & 50.9 & 34.2 & 22.7 & 19.7 & 41.2 & 28.8 & 40.3 & 34.0 \\
& MM-ECPE~\cite{ye2025mmecpe} & 73.4 & 72.3 & 76.8 & 57.5 & 41.7 & 28.9 & 24.7 & 49.5 & 65.2 & 49.2 & 66.7 \\
& \textbf{SAGML} & \textbf{80.2} & \textbf{79.6} & \textbf{82.7} & \textbf{62.4} & \textbf{46.0} & \textbf{31.8} & \textbf{26.9} & \textbf{52.7} & \textbf{71.8} & \textbf{53.8} & \textbf{73.4} \\
\midrule
\midrule
\multirow{9}{*}{EmVidCap}
& SA-LSTM~\cite{wang2018recnet} & 53.4 & 50.7 & 70.6 & 51.4 & 36.7 & 25.4 & 21.0 & 45.9 & 38.8 & 41.2 & 41.5 \\
& FT~\cite{wang2022emotion} & 51.2 & 49.6 & 67.6 & 47.2 & 32.0 & 21.6 & 20.4 & 43.1 & 29.0 & 37.6 & 33.3 \\
& SGN~\cite{ryu2021sgn} & 50.4 & 48.6 & 68.7 & 48.9 & 34.2 & 24.0 & 20.1 & 44.8 & 35.5 & 39.1 & 38.3 \\
& CANet~\cite{song2023contextual} & 53.7 & 52.7 & 68.1 & 47.7 & 32.9 & 22.5 & 19.7 & 43.7 & 34.5 & 38.8 & 38.2 \\
& VEIN~\cite{song2024vein} & 59.0 & 57.6 & 72.1 & 52.8 & 37.9 & 27.1 & 21.6 & 46.8 & 39.4 & 43.6 & 43.1 \\
& EPAN~\cite{song2023epan} & 69.3 & 67.2 & 74.4 & 55.6 & 39.9 & 28.0 & 23.0 & 47.1 & 43.0 & 47.0 & 48.0 \\
& DCGN~\cite{ye2024dcgn} & 74.8 & 73.1 & 75.6 & 56.7 & 40.5 & 28.5 & 24.9 & 51.7 & 49.8 & 48.5 & 51.7 \\
& MM-ECPE~\cite{ye2025mmecpe} & 75.6 & 73.8 & 78.1 & 58.5 & 42.3 & 30.2 & 26.4 & 53.8 & 67.9 & 50.4 & 69.3 \\
& \textbf{SAGML} & \textbf{82.3} & \textbf{81.5} & \textbf{85.7} & \textbf{64.3} & \textbf{46.1} & \textbf{34.8} & \textbf{28.5} & \textbf{57.2} & \textbf{76.9} & \textbf{55.7} & \textbf{77.9} \\
\bottomrule
\end{tabular}
\end{table*}

\subsection{Implementation Details}
For the default setting, we uniformly sample 30 frames and extract 512-dimensional CLIP ViT-B/32 features~\cite{radford2021clip}. A linear projection and a one-layer, one-head self-attention encoder map them to $D=300$. The graph contains 34 catalog nodes and 179 lexical nodes. Typed emotion strings are embedded with Qwen2.5-Instruct~\cite{qwen2025qwen25technicalreport}. The lexical adjacency combines Qwen similarity, 300-dimensional GloVe similarity~\cite{pennington2014glove}, and caption-corpus co-occurrence with weights $0.5$, $0.3$, and $0.2$, respectively, cosine edges below $\tau=0.5$ are removed. Add-$0.1$ smoothing is used for corpus-derived category--word and word--word statistics.

Visual and affective sequences are separately projected to the Qwen hidden size and concatenated as a continuous prefix. The Qwen backbone is adapted with LoRA~\cite{hu2022lora} on the query and value projections, using rank $8$, scaling factor $16$, and dropout $0.05$. Captions are truncated to 15 dataset words before being retokenized by the Qwen tokenizer. We train for 30 epochs with Adamax, learning rate $2\times10^{-4}$, weight decay $10^{-5}$, batch size 8, and gradient-norm clipping at 5.0. The emotion-distribution coefficient is $0.2$ for the first five epochs and $0.5$ thereafter. At inference, captions are generated autoregressively with greedy next-token selection. The feature study additionally uses ResNet-152 and ResNet-101+3D-ResNeXt-101 representations~\cite{he2016resnet,hara2018spatiotemporal}. All experiments were conducted on 8 NVIDIA Tesla V100 GPUs.

\subsection{Compared Methods}
We compare against two factual captioning baselines, SA-LSTM~\cite{wang2018recnet} and SGN~\cite{ryu2021sgn}, and seven EVC methods. FT uses separately trained fact and emotion streams~\cite{wang2022emotion}; CANet aggregates visual and textual context~\cite{song2023contextual}; VEIN learns an open-vocabulary visual emotion distribution~\cite{song2024vein}; EPAN introduces catalog-to-lexical tree-structured emotion learning~\cite{song2023epan}; DCGN evolves the emotion representation during decoding~\cite{ye2024dcgn}; MM-ECPE extracts paired emotions and visual causes~\cite{ye2025mmecpe}; and HEART aligns hierarchical visual semantics with emotion representations~\cite{han2026heart}. We transcribe the published results under the feature/decoder setting indicated by each source and use the three official test splits without resplitting.

\begin{figure*}[t]
    \centering
    \begin{center}
   \includegraphics[width=\linewidth]{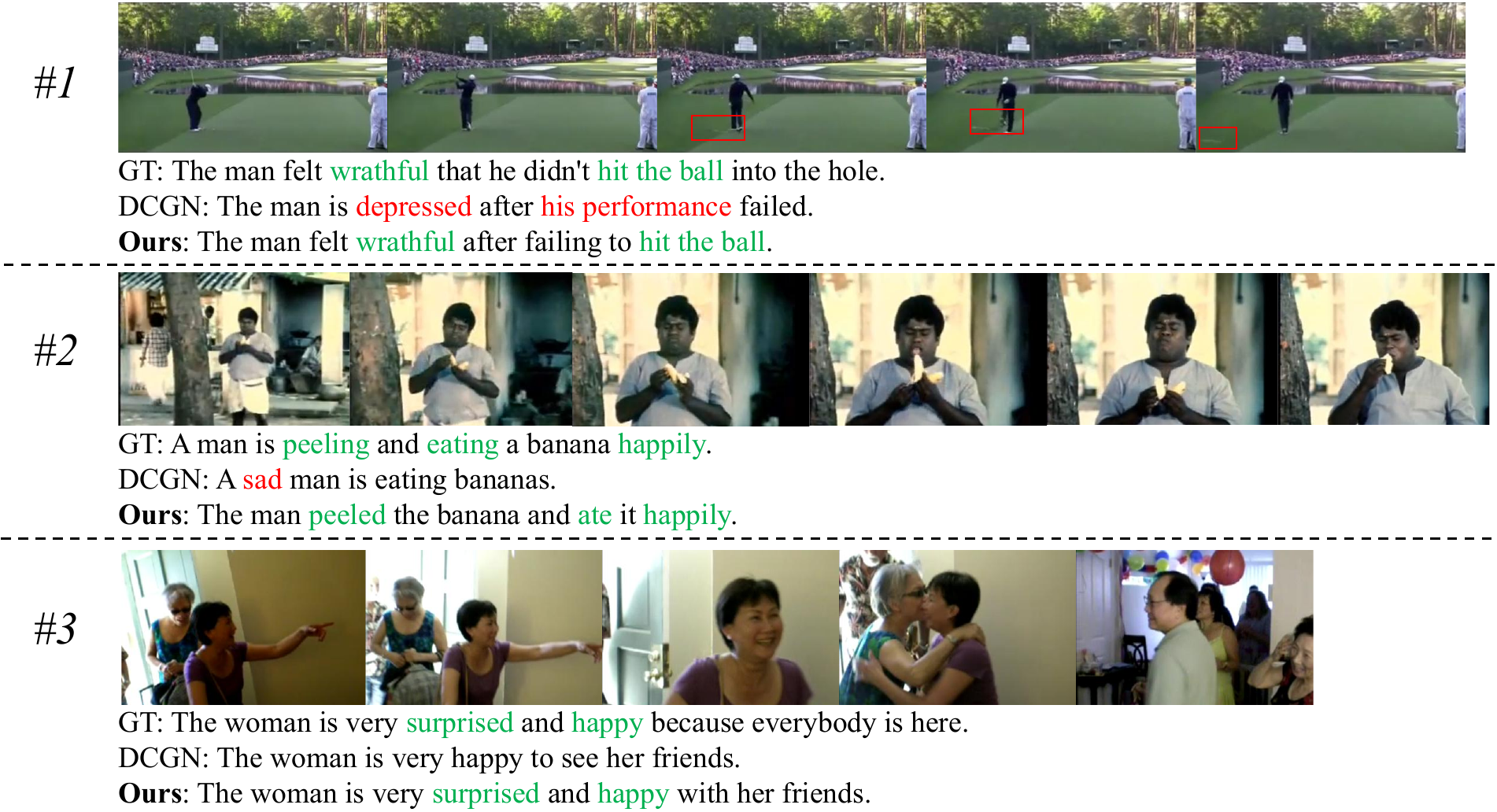}
    \end{center}
    \caption{Qualitative comparison between DCGN and SAGML on EmVidCap. SAGML more accurately associates emotions with visual events, preserves action details, and captures compound emotions. Green words indicate content consistent with the ground truth, while red words denote incorrect predictions.}
    \label{fig:case}
\end{figure*}

\begin{table*}[t]
\centering
\caption{Ablation study of core SAGML components on EmVidCap.}
\label{tab:ablation_components}
\begin{tabular}{lcc|ccccccc|cc}
\toprule
Method & Acc$_{sw}$ & Acc$_c$ & B-1 & B-2 & B-3 & B-4 & METEOR & ROUGE-L & CIDEr & BFS & CFS \\
\midrule
+ Catalog Emotion Perception & 74.3 & 72.1 & 81.7 & 59.4 & 42.5 & 31.6 & 25.2 & 52.3 & 62.8 & 51.0 & 64.9 \\
+ Soft Gate & 78.3 & 76.5 & 84.1 & 62.4 & 43.2 & 32.8 & 27.5 & 55.1 & 70.8 & 53.1 & 72.1 \\
+ Lexical Emotion Perception & 79.6 & 78.2 & 84.6 & 63.1 & 44.9 & 34.2 & 27.9 & 56.6 & 72.3 & 54.4 & 73.6 \\
SAGML full model & \textbf{82.3} & \textbf{81.5} & \textbf{85.7} & \textbf{64.3} & \textbf{46.1} & \textbf{34.8} & \textbf{28.5} & \textbf{57.2} & \textbf{76.9} & \textbf{55.7} & \textbf{77.9} \\
\bottomrule
\end{tabular}
\end{table*}

\begin{table*}[t]
\centering
\caption{Performance of SAGML with different visual features on EmVidCap.}
\label{tab:feature_comparison}
\begin{tabular}{lcc|ccccccc|cc}
\toprule
Feature & Acc$_{sw}$ & Acc$_c$ & B-1 & B-2 & B-3 & B-4 & METEOR & ROUGE-L & CIDEr & BFS & CFS \\
\midrule
ResNet-152 & 81.6 & 80.4 & 82.7 & 62.5 & 44.3 & 31.0 & 25.6 & 55.2 & 65.8 & 53.4 & 68.8 \\
ResNet-101 + 3D-ResNeXt-101 & 82.1 & 81.1 & 84.6 & \textbf{64.5} & 46.0 & 34.2 & \textbf{28.7} & \textbf{58.0} & 72.1 & 55.4 & 74.0 \\
CLIP & \textbf{82.3} & \textbf{81.5} & \textbf{85.7} & 64.3 & \textbf{46.1} & \textbf{34.8} & 28.5 & 57.2 & \textbf{76.9} & \textbf{55.7} & \textbf{77.9} \\
\bottomrule
\end{tabular}
\end{table*}

\begin{table*}[t]
\centering
\caption{Effect of the language decoder on EmVidCap.}
\label{tab:lm}

\begin{tabular}{lcc|ccccccc|cc}
\toprule
Method & Acc$_{sw}$ & Acc$_c$ & B-1 & B-2 & B-3 & B-4 & METEOR & ROUGE-L & CIDEr & BFS & CFS \\
\midrule
SAGML-BLIP & 80.5 & 78.9 & 82.8 & 60.3 & 40.5 & 31.9 & 26.8 & 54.7 & 71.4 & 52.1 & 73.1 \\
SAGML-Qwen & \textbf{82.3} & \textbf{81.5} & \textbf{85.7} & \textbf{64.3} & \textbf{46.1} & \textbf{34.8} & \textbf{28.5} & \textbf{57.2} & \textbf{76.9} & \textbf{55.7} & \textbf{77.9} \\
\bottomrule
\end{tabular}
\end{table*}

\subsection{Main Comparison With State-of-the-Art Methods}

Table~\ref{tab:main_evc} reports the comparison on all three benchmarks. SAGML obtains the highest value in every reported column, but the scale and source of the gains differ across the datasets.

\textbf{EmVidCap-S.} Compared with the strongest listed baseline, MM-ECPE, SAGML increases Acc$_{sw}$/Acc$_c$ from 88.7/87.6 to 93.5/92.3. The gains are not limited to emotion words: B-4 improves by 4.2 points, METEOR by 2.7, ROUGE-L by 2.6, and CIDEr by 7.8. CFS consequently rises from 88.3 to 95.5. Because this subset largely preserves MSVD factual content, the concurrent changes in semantic and emotion metrics indicate that SAGML is not merely appending frequent affective modifiers, the selected lexical emotions remain compatible with the reference event descriptions.

\textbf{EmVidCap-L.} This split is more difficult for every method because its clips and references are longer. SAGML reaches 80.2 Acc$_{sw}$ and 79.6 Acc$_c$, exceeding MM-ECPE by 6.8 and 7.3 points, respectively. It also improves CIDEr from 65.2 to 71.8 and CFS from 66.7 to 73.4. The improvement is consistent across B-1--B-4, METEOR, and ROUGE-L, suggesting that the soft graph prior remains useful when emotional expressions are less templated and must be coordinated with more factual words.

\textbf{EmVidCap.} On the combined split, SAGML improves the previous highest values by 6.7 points in Acc$_{sw}$, 7.7 in Acc$_c$, 4.6 in B-4, 9.0 in CIDEr, and 8.6 in CFS. The combined data contain both concise rewritten captions and longer free-form annotations, performance on this setting therefore reflects robustness to heterogeneous language patterns. The simultaneous increases in emotion, semantic, and hybrid measures support the intended role of the heterogeneous graph: catalog evidence guides lexical selection, while continuous gating leaves alternative emotion words available for visually conditioned attention and language decoding.

\subsection{Qualitative and Visualization Analysis}

Fig.~\ref{fig:case} compares captions generated by DCGN and SAGML for three representative videos. The examples examine complementary aspects of emotional video captioning: recognizing an emotion from an event outcome, preserving an action sequence while assigning the appropriate affect, and expressing multiple coexisting emotions. Green text marks content that agrees with the ground-truth caption, while red text marks an incorrect action or emotion.

In Example~\#1, the video shows a golfer missing a putt, as indicated by the ball trajectory in the highlighted frames. DCGN describes the man as ``depressed'' and refers only generally to a failed performance. SAGML instead generates ``wrathful'' and connects that emotion to the visually grounded event of failing to hit the ball, closely matching both the affective and factual content of the reference. This example illustrates the importance of linking an emotion word to its specific visual cause rather than relying on a broadly plausible negative emotion.

Example~\#2 requires both temporal action understanding and emotion recognition. The man first peels a banana and then eats it happily. DCGN captures only the eating action and incorrectly assigns the emotion ``sad''. SAGML preserves the two-step action sequence with ``peeled'' and ``ate'', while also predicting ``happily''. The caption therefore maintains the factual progression of the video without introducing an emotion that conflicts with its visual tone.

Example~\#3 contains compound affect. The reference describes the woman as both ``surprised'' and ``happy'' upon seeing the people around her. DCGN identifies happiness but omits surprise, reducing the event to a single dominant emotion. SAGML retains both emotions and grounds them in the woman's interaction with her friends. This result is consistent with the purpose of the soft heterogeneous graph: related lexical emotions can remain active simultaneously instead of being restricted to one coarse affective path.

Across the three cases, SAGML produces captions whose emotional expressions remain tied to observable people, actions, and event outcomes. The examples also complement the quantitative results: graph-based affective prompting helps select more appropriate and, when necessary, multiple emotion words, while the LLM decoder expresses that evidence in coherent factual sentences.

\subsection{Ablation Study}

We conduct ablations on EmVidCap to isolate the effects of catalog perception, topology-driven gating, lexical perception, graph relations, and prompt-level emotion distribution learning. Unless noted otherwise, all variants use the CLIP representation and Qwen decoder.

\subsubsection{Effectiveness of Main Components}
Table~\ref{tab:ablation_components} incrementally adds the main SAGML components. Catalog emotion perception alone provides a coarse affective prior but reaches only 74.3/72.1 on Acc$_{sw}$/Acc$_c$. Propagating this distribution into the topology-driven soft gate raises the two scores by 4.0 and 4.4 points and improves CIDEr by 8.0 and CFS by 7.2. This is the largest single change in the incremental comparison, showing that transferring catalog evidence to a non-binary lexical distribution is important for both affective choice and caption consensus.

Adding lexical emotion perception further improves B-4 from 32.8 to 34.2 and raises Acc$_c$ from 76.5 to 78.2. This stage allows frame tokens to query graph-refined lexical nodes rather than relying only on the propagated prior. Finally, the full model adds prompt-level emotion distribution learning and obtains 82.3/81.5 Acc$_{sw}$/Acc$_c$, 76.9 CIDEr, and 77.9 CFS. Relative to the preceding variant, the gains are larger on emotion accuracy and hybrid scores than on B-4, which is consistent with the objective directly constraining the affective content of the decoder prefix.

\subsubsection{Effect of Visual Representation}

To distinguish the contribution of SAGML from that of the input representation, we evaluate appearance-only ResNet-152 features, fused ResNet-101 and 3D-ResNeXt-101 appearance--motion features, and vision--language aligned CLIP features.

As shown in Table~\ref{tab:feature_comparison}, adding 3D motion features to the ResNet setting improves B-4 from 31.0 to 34.2, CIDEr from 65.8 to 72.1, and CFS from 68.8 to 74.0. Motion therefore contributes mainly to predicate and event description, while emotion accuracy changes more modestly. CLIP produces the highest Acc$_{sw}$, Acc$_c$, B-1, B-3, B-4, CIDEr, BFS, and CFS. Relative to the appearance--motion setting, its largest gains occur in CIDEr (4.8 points) and CFS (3.9 points), which is consistent with stronger visual--textual alignment. The motion-fused representation remains slightly higher on B-2, METEOR, and ROUGE-L. Thus, CLIP gives the best overall balance for SAGML, whereas explicit motion features retain an advantage on several overlap-oriented measures.

\subsubsection{Effect of Graph Relations}

\begin{table}[t]
\centering
\caption{Ablation of graph relation types on EmVidCap.}
\label{tab:ablation_graph}
\resizebox{\linewidth}{!}{
\begin{tabular}{lcccccc}
\toprule
Graph Variant & Acc$_{sw}$ & Acc$_c$ & B-4 & CIDEr & BFS & CFS\\
\midrule
Full heterogeneous graph & \textbf{82.3} & \textbf{81.5} & \textbf{34.8} & \textbf{76.9} & \textbf{55.7} & \textbf{77.9}\\
w/o $A_{cc}$ & 80.5 & 79.4 & 31.4 & 66.5 & 53.2 & 69.2\\
w/o $A_{ww}$ & 78.2 & 76.3 & 27.5 & 52.1 & 50.1 & 57.1\\
\bottomrule
\end{tabular}}
\end{table}

Table~\ref{tab:ablation_graph} examines the two within-level relation matrices while retaining category--word edges $A_{cw}$ in every variant. Removing catalog--catalog proximity $A_{cc}$ lowers Acc$_{sw}$/Acc$_c$ by 1.8/2.1 points and reduces CIDEr and CFS by 10.4 and 8.7, respectively. Neighboring catalog categories therefore provide useful uncertainty propagation beyond the initially predicted category.

The effect of lexical--lexical relations $A_{ww}$ is larger. Without them, B-4 decreases from 34.8 to 27.5, CIDEr from 76.9 to 52.1, and CFS from 77.9 to 57.1. The drop indicates that isolated category--word links are insufficient for selecting varied, semantically compatible emotion expressions. Combining $A_{cc}$ and $A_{ww}$ with the shared cross-layer matrix produces the most consistent result across all six measures.

\subsubsection{Effect of Emotion Distribution Learning}
Table~\ref{tab:ablation_edl} studies the two prompt-level emotion losses incrementally. Caption-only training already uses graph-derived affective tokens, but it does not directly constrain the pooled prompt state. Adding catalog EDL increases Acc$_{sw}$/Acc$_c$ from 79.6/78.2 to 81.5/79.4 and raises CIDEr by 2.9 points. Coarse distribution supervision therefore improves not only the predicted affect, but also the semantic use of the prefix.

Adding lexical EDL on top of catalog EDL further improves Acc$_c$ by 2.1 points, CIDEr by 1.7, and CFS by 1.6. The complete dual-level objective obtains the best value in every column. Catalog supervision stabilizes broad affective orientation, whereas lexical supervision more directly constrains the words that may appear in the caption. Their complementary effect is visible in the larger improvement of emotion and hybrid metrics.

\begin{table}[t]
\centering
\caption{Ablation of emotion distribution learning on EmVidCap.}
\label{tab:ablation_edl}
\resizebox{\linewidth}{!}{
\begin{tabular}{lcccccc}
\toprule
Objective & Acc$_{sw}$ & Acc$_c$ & B-4 & CIDEr & BFS & CFS\\
\midrule
Caption loss only & 79.6 & 78.2 & 34.2 & 72.3 & 54.4 & 73.6\\
+ Catalog EDL & 81.5 & 79.4 & 34.4 & 75.2 & 55.2 & 76.3\\
+ Lexical EDL & \textbf{82.3} & \textbf{81.5} & \textbf{34.8} & \textbf{76.9} & \textbf{55.7} & \textbf{77.9}\\
\bottomrule
\end{tabular}}
\end{table}

\subsection{Parameter Analysis}

\subsubsection{Effect of Lexical Similarity Threshold}
The lexical graph $A_{ww}$ retains Qwen- and GloVe-based similarity edges above threshold $\tau$. Table~\ref{tab:param_tau} compares $\tau\in\{0.4,0.5,0.6\}$. With $\tau=0.4$, the denser graph reaches 81.1 Acc$_{sw}$ but a lower CIDEr of 69.3, indicating that additional weak lexical edges introduce noise into caption generation. Increasing the threshold to 0.6 removes more associations and reduces all six measures, including a 4.5-point drop in B-4 and a 9.6-point drop in CFS relative to $\tau=0.5$. The intermediate value $\tau=0.5$ gives the best result throughout, balancing lexical connectivity against semantic selectivity.

\begin{table}[t]
\centering
\caption{Effect of lexical graph threshold $\tau$ on EmVidCap.}
\label{tab:param_tau}

\begin{tabular}{lcccccc}
\toprule
$\tau$ & Acc$_{sw}$ & Acc$_c$ & B-4 & CIDEr & BFS & CFS\\
\midrule
0.4 & 81.1 & 79.2 & 32.2 & 69.3 & 53.7 & 71.5\\
0.5 & \textbf{82.3} & \textbf{81.5} & \textbf{34.8} & \textbf{76.9} & \textbf{55.7} & \textbf{77.9} \\
0.6 & 79.5 & 77.2 & 30.3 & 65.8 & 51.8 & 68.3\\
\bottomrule
\end{tabular}
\end{table}

\subsection{Language Decoder Analysis}

The affective graph determines which emotion evidence is supplied to the generator, whereas the language decoder determines how that evidence is integrated with factual content. We compare a BLIP-based decoder~\cite{li2023blip2} with the Qwen-based decoder~\cite{qwen2025qwen25technicalreport} while retaining the SAGML visual encoder, graph construction, and emotion objectives.

Table~\ref{tab:lm} shows a consistent advantage for SAGML-Qwen across all metrics. Relative to SAGML-BLIP, it improves Acc$_{sw}$ by 1.8 points and Acc$_c$ by 2.6, indicating that the stronger decoder does not dilute the affective prefix. The semantic gains are also distributed across different matching criteria: B-1/B-2/B-3/B-4 increase by 2.9/4.0/5.6/2.9 points, METEOR by 1.7, ROUGE-L by 2.5, and CIDEr by 5.5.

The 5.6-point gain on B-3 and the 5.5-point gain on CIDEr are larger than the B-1 gain, suggesting that Qwen mainly improves multi-word composition and agreement with reference descriptions rather than only predicting additional isolated words. At the same time, BFS rises from 52.1 to 55.7 and CFS from 73.1 to 77.9. The joint movement of semantic, emotional, and hybrid metrics supports using Qwen as the default decoder: its linguistic prior better realizes the graph-selected affective evidence as coherent phrases, while the prompt-level catalog and lexical objectives keep generation tied to the video's emotion. Furthermore, the competitive performance under the BLIP setting also indicates that the capabilities of SAGML are not entirely dependent on LLM.

\section{Conclusion}

In this paper, we proposed SAGML, a self-adapting emotional video captioning framework that integrates affective heterogeneous graph reasoning with multi-task causal language modeling. Instead of using a rigid tree-structured emotion prior, SAGML represents catalog-level emotions and lexical-level emotion words as a unified heterogeneous graph, where category--category, category--word, and word--word relations are jointly modeled. To further connect structured affective reasoning with natural language generation, SAGML projects visual and graph-derived affective representations into the hidden space of a LoRA-adapted Qwen decoder. Prompt-level catalog and lexical emotion heads are introduced to supervise the decoder prefix, encouraging the language model to preserve multi-level emotional evidence while generating factually grounded captions. The overall framework is optimized with a joint objective that combines autoregressive caption generation and emotion distribution learning. Experiments on EmVidCap-S, EmVidCap-L, and the full EmVidCap benchmark demonstrate that SAGML consistently improves semantic, emotional, and hybrid evaluation metrics over existing emotional video captioning methods.


\bibliographystyle{IEEEtran}
\bibliography{references}

\vfill

\end{document}